\documentclass{article}
\usepackage{hyperref}
\hypersetup{
    colorlinks=true,
    linkcolor=blue,
    urlcolor=blue,
}

\usepackage{arxiv}
\renewcommand{\headeright}{}
\renewcommand{\undertitle}{}
\usepackage{placeins}
\usepackage[utf8]{inputenc} % allow utf-8 input
\usepackage[T1]{fontenc}    % use 8-bit T1 fonts
\usepackage{hyperref}       % hyperlinks
\usepackage{url}            % simple URL typesetting
\usepackage{booktabs}       % professional-quality tables
\usepackage{float}
\usepackage{amsfonts}       % blackboard math symbols
\usepackage{nicefrac}       % compact symbols for 1/2, etc.
\usepackage{microtype}      % microtypography
\usepackage{graphicx}
\usepackage{doi}

\usepackage[numbers]{natbib}
\title{Benchmarking LightGBM and BiLSTM for Sentiment Analysis on Indonesian E-Commerce Reviews}

\author{ 
    Lidia Natasyah Marpaung \\
    Department of Data Science \\
    Institut Teknologi Sumatera \\
    \texttt{lidia.123450015@student.itera.ac.id} \\
    \And
    Vania Claresta \\
    Department of Data Science \\
    Institut Teknologi Sumatera \\
    \texttt{vania.123450029@student.itera.ac.id} \\
    \And
    Iqfina Haula Halika \\
    Department of Data Science \\
    Institut Teknologi Sumatera \\
    \texttt{iqfina.123450076@student.itera.ac.id} \\
    \And
    Luluk Muthoharoh, M.Si \\
    Department of Data Science \\
    Institut Teknologi Sumatera \\
    \texttt{luluk.muthoharoh@sd.itera.ac.id} \\
    \And
    Ardika Satria, M.Si \\
    Department of Data Science \\
    Institut Teknologi Sumatera \\
    \texttt{ardika.satria@sd.itera.ac.id} \\
    \And
    Martin Clinton Tosima Manullang, Ph.D. \\
    Department of Informatics \\
    Institut Teknologi Sumatera \\
    \texttt{martin.manullang@if.itera.ac.id} \\
}

\renewcommand{\shorttitle}{Benchmarking LightGBM and BiLSTM}

\hypersetup{
pdftitle={Benchmarking LightGBM and BiLSTM for Sentiment Analysis on Indonesian E-Commerce Reviews},
pdfsubject={cs.CL, cs.LG},
pdfauthor={Lidia Natasyah Marpaung, Vania Claresta, Iqfina Haula Halika, Martin C.T. Manullang},
pdfkeywords={Sentiment Analysis, Machine Learning, Deep Learning, BiLSTM, PyCaret},
}

\begin{document}
\maketitle

\begin{abstract}
This study presents a comparative analysis between two primary approaches in Natural Language Processing (NLP): Machine Learning (ML) utilizing the PyCaret AutoML framework, and Deep Learning (DL). The evaluation is conducted on a sentiment analysis task using an Indonesian e-commerce review dataset sourced from Hugging Face. The dataset, consisting of 15,000 samples, is partitioned into training, validation, and testing sets. The ML experiments compare LightGBM, Logistic Regression, and Support Vector Machine (SVM) algorithms, whereas the DL experiment implements a Bidirectional Long Short-Term Memory (BiLSTM) architecture. The experimental results demonstrate that the BiLSTM model outperforms all ML models, achieving an accuracy of 98.87\% and an F1-Score of 98.87\%. Meanwhile, LightGBM emerges as the best-performing ML model with an accuracy of 98.23\% in a highly efficient training time. This research proves that the BiLSTM architecture is highly capable of capturing the sequential context of Indonesian review texts, making it the superior model for this specific classification task.

\end{abstract}


\section{Introduction}
The rapid growth of e-commerce in Indonesia has generated a massive volume of textual data in the form of product reviews. Sentiment analysis of these reviews is crucial for businesses to automatically understand customer satisfaction. Natural Language Processing (NLP) approaches have evolved from traditional Machine Learning (ML) modeling to more complex Deep Learning (DL) architectures. 

This research aims to benchmark the performance between an automated ML approach using PyCaret and a DL model for a text classification task. PyCaret offers the advantage of efficiently discovering the optimal ML algorithm automatically. Conversely, DL architectures, particularly Bidirectional Long Short-Term Memory (BiLSTM), are renowned for their exceptional ability to comprehend word sequence dependencies in both directions. The primary contribution of this experiment is to provide empirical evidence-based comparative guidance on which model delivers superior evaluation metrics on an Indonesian e-commerce dataset, as well as to provide model deployment results accessible via an interactive web interface.

\section{Related Work}
Sentiment analysis in Indonesian has been extensively explored in recent years. Several previous studies have focused on the use of standard classification algorithms coupled with TF-IDF based feature extraction to detect sentence polarity. For instance, studies employing Support Vector Machine (SVM) on Indonesian e-commerce and social media data have demonstrated strong classification performance. Alaiya and Agusniar \cite{Alaiya2025} applied SVM with a linear kernel on Tokopedia reviews, achieving an accuracy of 95.70\% using TF-IDF features with bigram configuration. Similarly, Singgalen \cite{Singgalen2024} evaluated SVM combined with the SMOTE oversampling technique for sentiment classification of YouTube tourism content, reaching an accuracy of 84.26\% and an AUC of 0.996. For simpler yet effective approaches, Fahrurrozi et al. \cite{Fahrurrozi2025} utilized Logistic Regression with TF-IDF on Indonesian reviews from Shopee, Tokopedia, and TikTok Shop, obtaining an overall accuracy of 88\%.

Beyond traditional classification algorithms, several studies have examined the role of feature extraction methods in sentiment analysis performance. Semary et al. \cite{Semary2024} conducted a comprehensive evaluation of various feature extraction techniques including Bag-of-Words (BOW), Word2Vec, N-gram, TF-IDF, Hashing Vectorizer, and GloVe on sentiment classification tasks. Their results demonstrated that TF-IDF achieved superior performance with 99\% accuracy on Amazon reviews and 96\% on Twitter airline data, supporting our choice of TF-IDF for the machine learning pipeline. Furthermore, Pongthao et al. \cite{Pongthao2025} compared multiple classification techniques including Logistic Regression, Decision Tree, Random Forest, and Support Vector Machine on health datasets, finding that SVM consistently outperformed other models with accuracies ranging from 73\% to 76\%, which aligns with the strong baseline performance of SVM in our study.

However, the application of Deep Learning in text processing indicates a paradigm shift. Recurrent Neural Network (RNN) based architectures, specifically LSTM and its variations (BiLSTM), have been proven to significantly improve classification precision on texts containing slang and unstructured sentences, which are typical of social media and e-commerce platforms \cite{Cai2020}. In the context of Indonesian language, Farizi and Sibaroni \cite{Farizi2025} combined BiLSTM with IndoBERT for sentiment analysis of TikTok reviews, achieving 81\% accuracy on the classification report test and 92.03\% on 10-fold cross-validation. Anadra et al. \cite{Anadra2025} further compared BiLSTM and IndoBERT on Tokopedia reviews, finding that IndoBERT achieved higher predictive accuracy (balanced accuracy up to 0.85), while BiLSTM offered greater training efficiency. Nasution et al. \cite{Nasution2025} conducted a comparative study between BiLSTM and GRU on Indonesian e-commerce product reviews, where GRU with SGD optimizer achieved the highest performance at 94.07\% accuracy, while BiLSTM with SGD yielded a competitive 93.61\% accuracy.

\subsection{BiLSTM Applications in Indonesian Context}
The effectiveness of BiLSTM for Indonesian sentiment analysis has been validated across multiple domains. Setiawan et al. \cite{Setiawan2023} compared LSTM with IndoBERTweet for TikTok review classification, demonstrating that IndoBERTweet achieved 80\% accuracy while LSTM reached 78\%. Haspin et al. \cite{Haspin2025} employed BiLSTM with Word2Vec embeddings for TikTok opinion mining, achieving 80\% accuracy and an F1-score of 0.78, though neutral sentiments proved challenging to detect. Romli et al. \cite{Romli2022} compared CNN and BiLSTM for sentiment and emotion analysis of Indonesian COVID-19 related tweets, finding BiLSTM superior with 69.48\% accuracy for sentiment analysis and 84.36\% for emotion analysis, outperforming CNN's 68.58\% and 84.21\% respectively. Ningrum et al. \cite{Ningrum2025} developed an automated comment bot system using BERT for TikTok sentiment analysis targeting Indonesian SMEs, achieving an impressive 91.94\% accuracy.

\subsection{Cross-Domain Applications}
Beyond e-commerce, sentiment analysis techniques have been applied to diverse domains. Cai et al. \cite{Cai2020} proposed a BERT-BiLSTM combination model for analyzing investor and consumer sentiment in energy markets, achieving 0.8620 accuracy and 0.7078 recall, demonstrating the effectiveness of hybrid approaches. Hidayat et al. \cite{Hidayat2021} utilized Doc2Vec with SVM and Logistic Regression for sentiment analysis of Twitter data related to environmental development, achieving accuracies above 75\%. Rana et al. \cite{Rana2023} developed a BiLSTM-CF and BiGRU-based deep sentiment analysis model for customer review recommendations, showcasing the versatility of bidirectional architectures in recommendation systems.

\subsection{Product-Specific Sentiment Analysis}
Recent work has also focused on specific product categories. Purnasiwi \cite{Purnasiwi2023} investigated sentiment analysis on skincare product reviews using Word Embedding and LSTM methods. Fauziah et al. \cite{Fauziah2025} analyzed TikTok comments on Skintific skincare products, while Nurjanah and Apidana \cite{Nurjanah2025} examined TikTok sentiment to evaluate brand reputation post-controversy for Daviena Skincare, demonstrating the growing importance of social media sentiment analysis for brand management in the Indonesian market.

These diverse studies collectively demonstrate the evolution from traditional ML approaches to sophisticated deep learning architectures for Indonesian sentiment analysis, with BiLSTM emerging as a particularly effective architecture for capturing the nuances of informal Indonesian language in e-commerce contexts.

\section{Dataset}
The quality and characteristics of the data play a pivotal role in the success of any natural language processing task. This section details the data acquisition, underlying characteristics, and the rigorous preparation steps undertaken before modeling.

\subsection{Data Acquisition and Sampling}
\begin{figure}[htbp]
    \centering
    \includegraphics[width=0.6\textwidth]{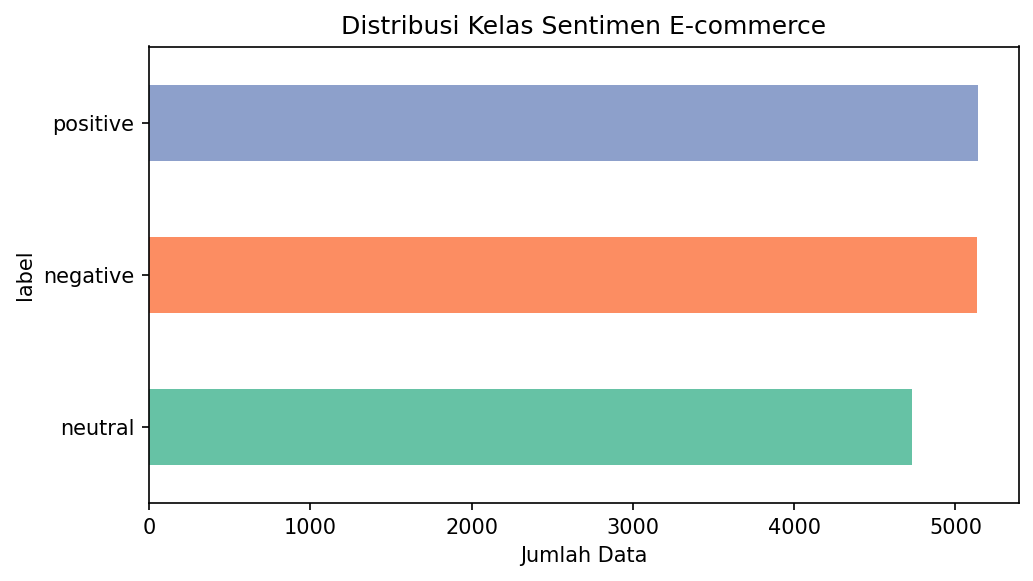}
    \caption{Label distribution of the e-commerce sentiment dataset.}
    \label{fig:label_dist}
\end{figure}

The primary dataset utilized in this study is sourced from the open-source Hugging Face repository (\texttt{AIbnuHibban/e-commerce-sentiment-bahasa-indonesia}). The original corpus contains 21,840 rows of user reviews collected from various Indonesian e-commerce platforms. However, to optimize computational efficiency and align with the strictly defined parameter thresholds of the Deep Learning architecture, a representative subset of 15,000 rows was extracted. As illustrated in Figure \ref{fig:label_dist}, the sampled dataset maintains a well-balanced distribution across three target classes: Positive, Negative, and Neutral. This balanced nature is crucial to prevent the models from developing a majority-class bias during the training phase.

\subsection{Data Characteristics and Linguistic Challenges}
Indonesian e-commerce reviews are notoriously complex due to the heavy use of informal language, code-mixing, and implicit emotional expressions. Table \ref{tab:dataset_sample} provides a snapshot of the dataset, illustrating the intricate linguistic challenges the models must navigate. 

\begin{table}[htbp]
    \caption{Sample rows from the e-commerce sentiment dataset}
    \centering
    \begin{tabular}{p{7.5cm}lcc}
        \toprule
        \textbf{Comment} & \textbf{Sentiment} & \textbf{Category} & \textbf{Rating} \\
        \midrule
        Wah mantap banget nih ditipu! Terima kasih seller & negative & sarcasm & 1 \\
        Bagus sekali, barang rusak semua! Top markotop! & negative & sarcasm & 1 \\
        Pelayanan terbaik! Nungguin 2 minggu cuma buat barang ga dateng & negative & sarcasm & 1 \\
        Seneng banget barangnya cacat. Recommended seller! & negative & sarcasm & 1 \\
        Keren nih seller! Kirim barang yang salah dengan percaya diri & negative & sarcasm & 1 \\
        
        \multicolumn{4}{c}{\vspace{0.2cm} \dots \vspace{0.2cm}} \\ 
        
        Lumayan bagus, tapi masih banyak kompetitor yang lebih baik. & neutral & - & 3 \\
        Mengecewakan! Barang tidak sesuai spesifikasi yang dijanjikan. & negative & - & 2 \\
        Jelek! Ukuran tidak pas dan finishing berantakan. & negative & - & 2 \\
        Gimana ya sih, ga sesuai warna & negative & - & 1 \\
        Produknya standar, sesuai harga yang dibayar. & neutral & - & 3 \\
        \bottomrule
    \end{tabular}
    \label{tab:dataset_sample}
\end{table}

Notably, the head of the dataset highlights instances of sarcasm. In these cases, inherently negative experiences are masked with highly positive vocabulary (e.g., "Wah mantap banget nih ditipu!"). These sarcastic structures pose a significant hurdle for traditional Bag-of-Words or TF-IDF approaches, as the models must interpret the sequence and context rather than just isolated words. Conversely, the tail of the table shows more straightforward negative and neutral evaluations, often accompanied by corresponding numerical ratings. The detection of sarcasm and implicit sentiment in Indonesian reviews poses significant challenges, as noted by Anadra et al. \cite{Anadra2025}, who found that preprocessing methods and labeling strategies significantly impact model performance on such nuanced expressions. The use of Word2Vec and other embedding techniques, as explored by Romli et al. \cite{Romli2022} and Purnasiwi \cite{Purnasiwi2023}, can help models better capture the semantic relationships necessary for understanding context-dependent sentiment.

\subsection{Data Partitioning and Preprocessing}
To ensure a robust and unbiased evaluation, the sampled data was partitioned into three distinct subsets using a stratified split: 12,000 rows (80\%) were allocated for the training set, 1,500 rows (10\%) for the validation set, and 1,500 rows (10\%) for the final testing set. 

Furthermore, a rigorous text preprocessing pipeline was implemented to standardize the corpus and eliminate structural noise. This pipeline encompassed four sequential operations:
\begin{enumerate}
    \item \textbf{Case Folding:} Lowercasing all characters to ensure uniformity (e.g., treating "Bagus" and "bagus" as the same token).
    \item \textbf{Noise Removal:} Stripping URLs, HTML tags, and metadata that hold no semantic value.
    \item \textbf{Character Filtering:} Eliminating non-alphanumeric characters, excessive punctuation, and irrelevant symbols.
    \item \textbf{Slang Normalization:} Mapping common, abbreviated Indonesian slang words heavily used in online marketplaces (e.g., "yg", "bgus", "gk") into their standard vocabulary forms. This final step is particularly vital for reducing the out-of-vocabulary (OOV) rate and improving the embedding quality in the neural network pipeline.
\end{enumerate}

\section{Methodology}
The research methodology is structured into three primary stages: data preprocessing, the Machine Learning experimental pipeline, and the Deep Learning experimental pipeline. This comprehensive approach is designed to systematically benchmark traditional algorithms against neural network architectures for sentiment classification.

\subsection{Data Preprocessing}
Before feeding the e-commerce reviews into the modeling pipelines, rigorous text preprocessing was conducted to ensure data quality and reduce noise. The preprocessing steps were implemented utilizing standard Python libraries, incorporating regular expressions (\texttt{re}) and \texttt{NLTK}. The pipeline included lowercasing all text, alongside the removal of punctuation, URLs, special characters, and numbers. Given the nature of Indonesian e-commerce reviews, specific attention was given to normalizing slang words and handling emojis that often carry significant sentiment weight. Finally, tokenization and stopword removal were applied to filter out structurally necessary but semantically empty words, leaving only the core tokens required for effective sentiment extraction.

\subsection{Machine Learning Pipeline (PyCaret)}
The first experimental pipeline utilizes the \texttt{pycaret.classification} module to automate the training, tuning, and evaluation of traditional Machine Learning models. Prior to model training, the cleaned text data was transformed into numerical vectors using Term Frequency-Inverse Document Frequency (TF-IDF) feature extraction, capped at a maximum of 5,000 features to maintain computational efficiency. This vectorization technique is crucial for capturing the relative importance of words within the dataset while penalizing overly frequent terms that do not contribute to sentiment polarity.

The choice of TF-IDF as our feature extraction method is well-supported by prior research. Semary et al. \cite{Semary2024} demonstrated that TF-IDF consistently outperforms other feature extraction techniques including BOW, Word2Vec, and GloVe across multiple datasets, achieving up to 99\% accuracy. This aligns with findings from Alaiya and Agusniar \cite{Alaiya2025} who successfully employed TF-IDF with bigram configuration (max\_features=5000, ngram\_range=(1,2)) for Indonesian e-commerce sentiment analysis.

Within the PyCaret environment, the automated pipeline concurrently benchmarked three primary algorithms: Light Gradient Boosting Machine (LightGBM), Logistic Regression, and Support Vector Machine (SVM) configured with a Linear Kernel. To ensure the robustness of the models and avoid overfitting, a stratified 10-fold cross-validation strategy was implemented. Furthermore, PyCaret's automated hyperparameter tuning was leveraged to find the optimal configuration for each algorithm, systematically exploring the search space for parameters such as learning rate, regularization strength, and maximum tree depth, while optimizing for the F1-Score.

\subsection{Deep Learning Architecture (BiLSTM)}
The second pipeline employs a neural network architecture developed entirely within the PyTorch framework. The core of this model is a Bidirectional Long Short-Term Memory (BiLSTM) network. Unlike standard LSTMs, the BiLSTM architecture processes the input text sequences in both forward and backward directions simultaneously, which is highly effective in capturing long-term dependencies and contextual nuances such as sarcasm and negation. The BiLSTM architecture's bidirectional processing capability makes it particularly effective for sentiment analysis tasks where context from both directions is crucial. This has been validated in multiple studies: Romli et al. \cite{Romli2022} found BiLSTM superior to CNN for Indonesian sentiment analysis, while Haspin et al. \cite{Haspin2025} demonstrated its effectiveness with Word2Vec embeddings, and Cai et al. \cite{Cai2020} showed strong results when combining BiLSTM with BERT for complex sentiment detection tasks.

The network architecture is initialized with an embedding layer of dimension 128, which converts tokenized words into dense vector representations. These embeddings are subsequently fed into the BiLSTM layer configured with a hidden state size of 64. To mitigate the risk of overfitting during the training phase, spatial dropout layers with a rate of 0.3 were integrated into the architecture. Finally, the extracted sequential features are passed through fully connected (Dense) layers. The final output layer utilizes a Softmax activation function to compute the prediction probabilities for the three distinct sentiment classes (Positive, Negative, and Neutral).

The model was trained in mini-batches of 32 over a maximum of 20 epochs. Optimization was handled by the Adam optimizer with an initial learning rate of $1 \times 10^{-3}$, utilizing Cross-Entropy Loss as the objective function. In strict accordance with the experimental constraints, the network was meticulously engineered to maintain computational efficiency, ensuring the total parameter count remained at approximately 1.5 million, well below the 10 million parameter threshold. The training process was monitored closely, yielding an average execution time of approximately 3.7 minutes.

\begin{figure}[htbp]
    \centering
    \includegraphics[width=0.6\textwidth]{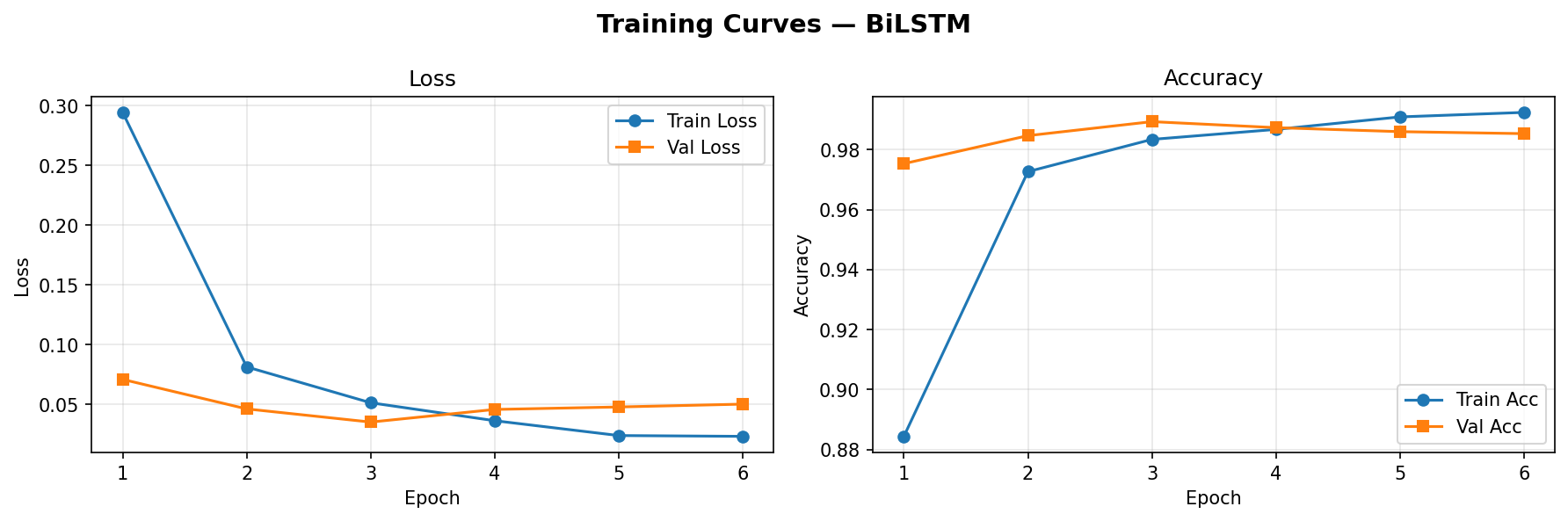}
    \caption{Training and validation curves (Loss and Accuracy) for the BiLSTM model.}
    \label{fig:train_curves}
\end{figure}

\section{Results \& Discussion}
The evaluation phase of this study revealed highly competitive performance metrics across both the traditional Machine Learning and Deep Learning pipelines. The following subsections detail the quantitative results and provide a comparative analysis of the employed architectures.

\subsection{Machine Learning Models Evaluation}
Based on the initial automated benchmarking conducted via PyCaret, tree-based ensemble methods significantly outperformed linear models. LightGBM emerged as the superior algorithm, achieving near-perfect evaluation metrics with an accuracy, precision, recall, and F1-score uniformly recorded at 0.9823. Furthermore, LightGBM demonstrated exceptional computational efficiency, requiring only 4.932 seconds for the entire training and cross-validation process. 

As presented in Table \ref{tab:ml_benchmark}, Logistic Regression and Support Vector Machine (SVM) with a Linear Kernel also established very strong baselines, achieving accuracies of 0.9767 and 0.9756, respectively. However, their slightly lower performance compared to LightGBM suggests that while the TF-IDF feature space is linearly separable to a large extent, the gradient boosting framework of LightGBM is more adept at capturing complex, non-linear relationships within the extracted text features.

\begin{table}[htbp]
    \caption{Machine Learning Models Benchmark (PyCaret)}
    \centering
    \begin{tabular}{lccccccc}
        \toprule
        \textbf{Model} & \textbf{Accuracy} & \textbf{AUC} & \textbf{Recall} & \textbf{Precision} & \textbf{F1-Score} & \textbf{Kappa} & \textbf{Time (sec)} \\
        \midrule
        LightGBM & 0.9823 & 0.9984 & 0.9823 & 0.9823 & 0.9823 & 0.9734 & 4.932 \\
        Logistic Regression & 0.9767 & 0.0000 & 0.9767 & 0.9768 & 0.9767 & 0.9650 & 9.162 \\
        SVM - Linear Kernel & 0.9756 & 0.0000 & 0.9756 & 0.9758 & 0.9756 & 0.9633 & 10.952 \\
        \bottomrule
    \end{tabular}
    \label{tab:ml_benchmark}
\end{table}

\subsection{Deep Learning Model Evaluation (BiLSTM)}
Conversely, the Deep Learning approach yielded an even more refined classification capability. The BiLSTM model recorded the highest overall performance, achieving a test accuracy of 98.87\%. A detailed breakdown of the classification report is provided in Table \ref{tab:bilstm_report}. 

The results indicate that the BiLSTM architecture successfully managed the sentiment classification task with remarkable consistency across all categories. The model classified both Negative and Neutral classes with an outstanding precision and recall of 0.99. Meanwhile, the Positive class achieved a precision of 0.98 and a recall of 0.99. The balanced support values (513 Negative, 473 Neutral, and 514 Positive) confirm that these high metric scores are robust and not the result of a heavily imbalanced dataset bias.

\begin{table}[htbp]
    \caption{Classification Report for BiLSTM Model}
    \centering
    \begin{tabular}{lcccc}
        \toprule
        \textbf{Class} & \textbf{Precision} & \textbf{Recall} & \textbf{F1-Score} & \textbf{Support} \\
        \midrule
        Negative & 0.99 & 0.99 & 0.99 & 513 \\
        Neutral  & 0.99 & 0.99 & 0.99 & 473 \\
        Positive & 0.98 & 0.99 & 0.99 & 514 \\
        \bottomrule
    \end{tabular}
    \label{tab:bilstm_report}
\end{table}

\subsection{Confusion Matrix and Error Analysis}
\begin{figure}[htbp]
    \centering
    \includegraphics[width=0.5\textwidth]{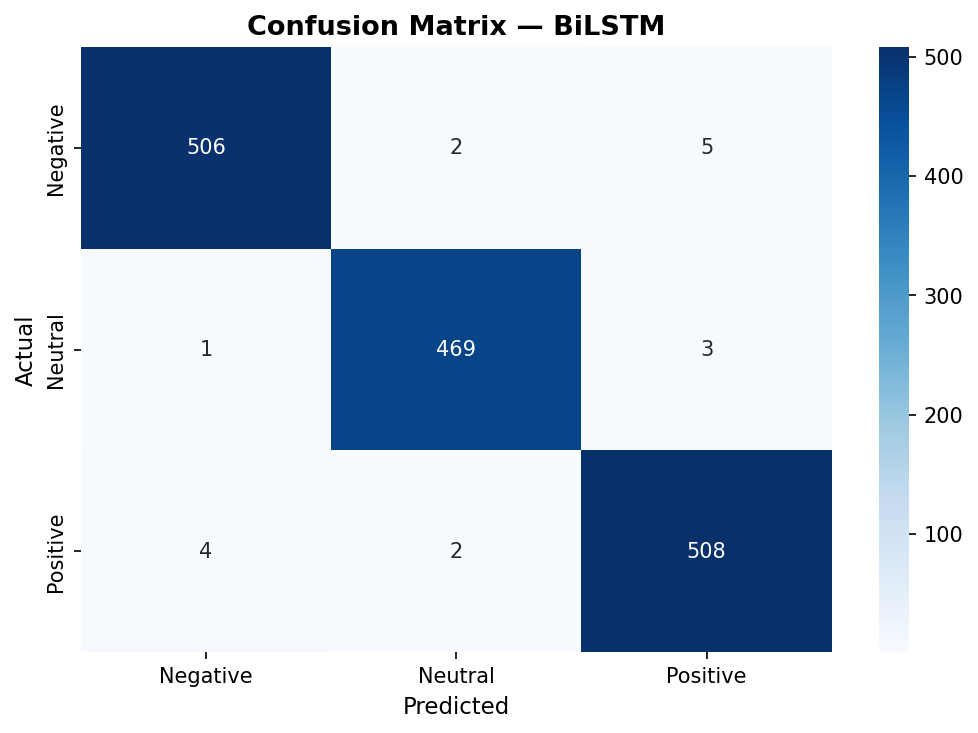}
    \caption{Confusion Matrix for the BiLSTM Model.}
    \label{fig:conf_matrix}
\end{figure}

To gain a deeper understanding of the BiLSTM model's predictive behavior and misclassification patterns, a confusion matrix was generated (Figure \ref{fig:conf_matrix}). The matrix visually corroborates the quantitative metrics, showing a heavy concentration of predictions along the main diagonal, which represents true positives for each class. The minimal off-diagonal elements indicate that the model rarely confused diametrically opposed sentiments (e.g., misclassifying a strongly negative review as positive). The few existing errors are likely attributed to highly ambiguous phrasing, implicit sarcasm, or mixed sentiments within a single review that still pose a challenge to sequence models.

\FloatBarrier
\subsection{Comparative Discussion}
The training curves previously illustrated in Figure \ref{fig:train_curves} demonstrated a steady convergence of loss and an increase in accuracy without significant overfitting, proving the stability of the BiLSTM training process. 

When comparing the two pipelines, a distinct trade-off between computational efficiency and predictive nuance is evident. The automated LightGBM model offers a highly scalable solution, achieving 98.23\% accuracy in merely 5 seconds. In contrast, the BiLSTM model requires significantly more computational resources and a longer training time (approximately 3.7 minutes). However, its bidirectional mechanism allows it to capture the sequential and contextual semantic dependencies of the Indonesian language more effectively, ultimately yielding a superior accuracy of 98.87\%. For e-commerce platforms where maximum precision in understanding customer feedback is prioritized over instantaneous model retraining, the BiLSTM architecture proves to be the more optimal solution.

Our findings align with broader trends in sentiment analysis research. The trade-off between computational efficiency and predictive accuracy has been well-documented: Anadra et al. \cite{Anadra2025} noted that BiLSTM requires only 1-2.5 minutes per epoch compared to IndoBERT's 2.6-3.6 minutes, while Pongthao et al. \cite{Pongthao2025} demonstrated that SVM offers excellent accuracy with minimal computational overhead across multiple classification tasks. The choice between ML and DL approaches ultimately depends on deployment context, with ML models like LightGBM offering superior speed (4.93 seconds in our study) for production environments requiring frequent retraining, while BiLSTM provides the highest accuracy (98.87\%) for applications where prediction quality is paramount.

\FloatBarrier
\section{Conclusion}
Based on the experimental results and comparative evaluation, the Bidirectional Long Short-Term Memory (BiLSTM) Deep Learning model is determined to be the optimal classification model for this Indonesian e-commerce sentiment dataset. BiLSTM achieved a peak accuracy of 98.87\%, surpassing the best Machine Learning algorithm, LightGBM, which scored 98.23\%. The primary advantage of the BiLSTM lies in its ability to consistently maintain a balance between precision and recall across all three sentiment classes, facilitated by its bidirectional sequential learning mechanism. Although it requires a longer training duration, approximately 3.7 minutes compared to LightGBM's 4.9 seconds, the margin of accuracy and the stability of the loss make BiLSTM a more robust solution for detecting the polarity of reviews characterized by heterogeneous language.

\section*{Data and Code Availability}
The complete source code, dataset, and links to the deployed Hugging Face models are available on GitHub at \url{https://github.com/LidiaNatasyah/pba2026-Kelompok11}.
 
\bibliography{references}
\end{document}